# Deep Transfer Learning for Industrial Automation

A Review and Discussion of New Techniques for Data-Driven Machine Learning

Benjamin Maschler[1] and Michael Weyrich[2]

**ABSTRACT** In this article, the concepts of transfer and continual learning are introduced. The ensuing review reveals promising approaches for industrial deep transfer learning, utilizing methods of both classes of algorithms. In the field of computer vision, it is already state-of-the-art. In others, e.g. fault prediction, it is barely starting. However, over all fields, the abstract differentiation between continual and transfer learning is not benefitting their practical use. In contrast, both should be brought together to create robust learning algorithms fulfilling the industrial automation sector's requirements. To better describe these requirements, base use cases of industrial transfer learning are introduced.

## 1. INTRODUCTION

Deep learning has greatly increased the capabilities of 'intelligent' technical systems over the last years [1]. This includes the industrial automation sector [1–4], where new data-driven approaches to e.g. predictive maintenance [2], computer vision [3] or anomaly detection [4] resulted in systems more easily and robustly automated than ever before.

However, the practical implementation of those advances is obstructed by two characteristics of deep learning [2, 5–7]:

Firstly, training datasets and actual problem need to be very similar regarding their feature space and the distribution of data therein. Furthermore, only effects that are represented within the training data can be learned by the algorithm. This results in training datasets needing to be very large and diverse in order to include rare events as well. In practice, such datasets are increasingly difficult to obtain as problems become more complex [2, 5, 6].

Secondly, retraining a deep learning algorithm once trained is not much different from training a completely untrained deep learning algorithm. Both require vast amounts of computational power and all training data. In the highly dynamic industrial automation environment, where e.g. one production line might switch products, tools or processes regularly, this is impractical [5, 7].

Yet, both problems can be mitigated by *'transfer learning'*, i.e. a set of approaches aiming to reduce the amount and quality of data needed as well as providing a way to build upon previously acquired knowledge as opposed to starting every learning process from scratch [5, 7]. This is achieved by transferring knowledge among tasks creating distributed cooperative learning systems.

[1] University of Stuttgart, Institute of Industrial Automation and Software Engineering, Pfaffenwaldring 47, 70569 Stuttgart, Germany, +49 711 685 67295, benjamin.maschler@ias.uni-stuttgart.de, ORCID: 0000-0001-6539-3173
[2] University of Stuttgart, Institute of Industrial Automation and Software Engineering

Consequently, this article introduces the different concepts related to transfer learning in general (section 2). From there, it examines potentials as well as the current usages of deep learning based transfer learning from a use case oriented perspective for the field of industrial automation (section 3) in order to give directions to readers interested in adapting it to their own needs (section 4).

## 2. BRIEF REVIEW OF TRANSFER LEARNING

The term *transfer learning* was first discussed on a broader scale regarding machine learning during a 1995 workshop on "Learning to Learn" addressing the long-term goal of lifelong machine learning. Similar to 'natural' intelligence in higher mammals, this would enable machines' learning to build upon previously acquired knowledge. This definition was subsequently further narrowed down to the transfer of knowledge and skills from previously learned tasks to novel tasks by the Defense Advanced Research Projects Agency (DARPA) in 2005 [8]. It concentrates solely on the novel '*target task*', without regards to the algorithms continued capability to potentially solve old '*source tasks*' as well. Algorithms aimed at solving target as well as source tasks using previously acquired knowledge where thereon called '*multi-tasking*' or '*continual learning*' algorithms [8] (see Figure 1).

However, the authors believe that in today's industrial automation use cases, both objectives closely coexist. Therefore, both fields are briefly presented and systematically brought together in the following, laying the grounds for the subsequent evaluation regarding their use in industrial automation. In order to ease the access to those concepts, important terms, their definitions and a coherent example taken from the optical detection of manufacturing defects on printed circuit assemblies are given in Table 1.

### 2.1 TRANSFER LEARNING: TERMS AND DEFINITIONS

The field of transfer learning examines and develops machine learning methods using knowledge acquired from previously solved *source tasks* in order to more efficiently solve new *target tasks*. According to [8–10], it can be structured by analyzing the problems it solves or the approaches it uses to do so.

*Problem categorization* aims at defining different categories of transfer learning based upon characterizing the types of problems they solve. More specifically, it either looks at the availability of labeled data from the source or target domains or on the similarity of source and target (input) feature spaces.

Regarding the availability of labeled data, three different categories are commonly distinguished:

- **Inductive Transfer Learning** describes settings in which target domain labels are available [8–10]. Depending on the additional availability of source domain labels, this category has similarities with *multi-tasking* (if source domain labels are available) [11, 12] or *self-taught learning* (if source domain labels are not available) [13].
- **Transductive Transfer Learning** describes settings in which only source domain labels are available [8–10].
- **Unsupervised Transfer Learning** describes settings in which neither source nor target domain labels are available [8–10].

Regarding the similarity of (input) feature spaces, two categories are commonly distinguished:

- **Homogenous Transfer Learning** describes settings in which source and target feature spaces are identical [8, 9].
- **Heterogenous Transfer Learning** describes settings in which source and target feature spaces differ [8, 9].

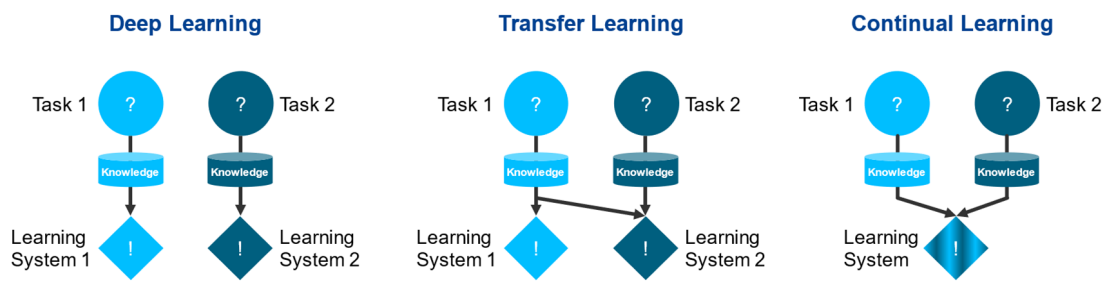

FIGURE 1. **Differences in the learning process between conventional deep learning, transfer learning and continual learning**

**TABLE 1.** Important terms, their definitions and a coherent example linking them

| *Terms* | *Definitions* | *Examples* |
|---|---|---|
| Label | Desired output of the algorithm | 'Defect' |
| Label Space | Space containing all possible labels | {'OK', 'Defect'} |
| Feature | Individual characteristic of the input of the algorithm | 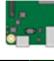 |
| Feature Space | Space containing all possible features | { 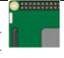 , 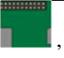 , … , 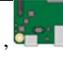 , 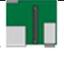 } |
| Instance | Set of features and input of the algorithm | 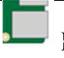 |
| Decision function | Function returning correct label for every instance | Algorithm that can determine a products label based upon the features of that products image. |
| Task | Characterized by label space and (to be learned) decision function | Determine a product's label based upon an image of that product. |
| Dataset | Specific set of (labeled or un-labeled) instances | {( 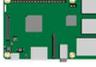 , 'OK'), ( 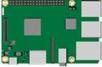 , 'Defect'), ( 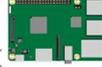 , 'Defect'), ( 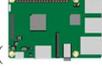 , 'Defect')} |
| Domain | Defined by feature space and marginal probability distribution of instances | Scenario comprising all possible products' images and their specific incidences. |

In literature, sometimes the same categories are applied to the similarity of (output) label spaces as well (see e.g. [9, 14]). However, in this article no such differentiation is made, because no clear terminological trends could be discerned by the authors.

Although being widely used as stated above, there still are inconsistencies in the terminology regarding transfer learning. Yet, throughout this article, only the definitions formulated in this section will be used. A more detailed discussion on the topic of transfer learning terminology can be found in [9].

*Solution categorization* aims at defining different categories of transfer learning based upon characterizing the types of approaches they use. Generally, a first differentiation is made between *statistical transfer learning* and *deep transfer learning* approaches, the latter comprising of approaches using deep neural networks to realize machine learning [10, 15]. Yet, both can be divided into four main approach categories:

- **Instance Transfer** (see Figure 2, Case A) describes approaches that add (weighted) instances from the source domain(s) to the target domain to improve training on the target task [8–10]. This increases the amount of data available for training without substantially changing the algorithm itself. It can therefore be applied in the same way to statistical as well as deep transfer learning [15].
- **Feature Representation Transfer** (see Figure 2, Case B) describes approaches that map instances from the source and target domains into a common feature space to improve training on the target task [8–10]. In deep transfer learning, this can be achieved using feature extractors for domain adaption prior to the decision function [14, 15].
- **Parameter Transfer** (see Figure 2, Case C) describes approaches that share parameters or priors between the source and target domain models improving the initial model before training on the target task itself begins [8–10]. This changes the algorithm itself, lowering its need for training data. For deep transfer learning, this corresponds to a partial re-use of deep neural networks pre-trained on the source domain(s) [15].

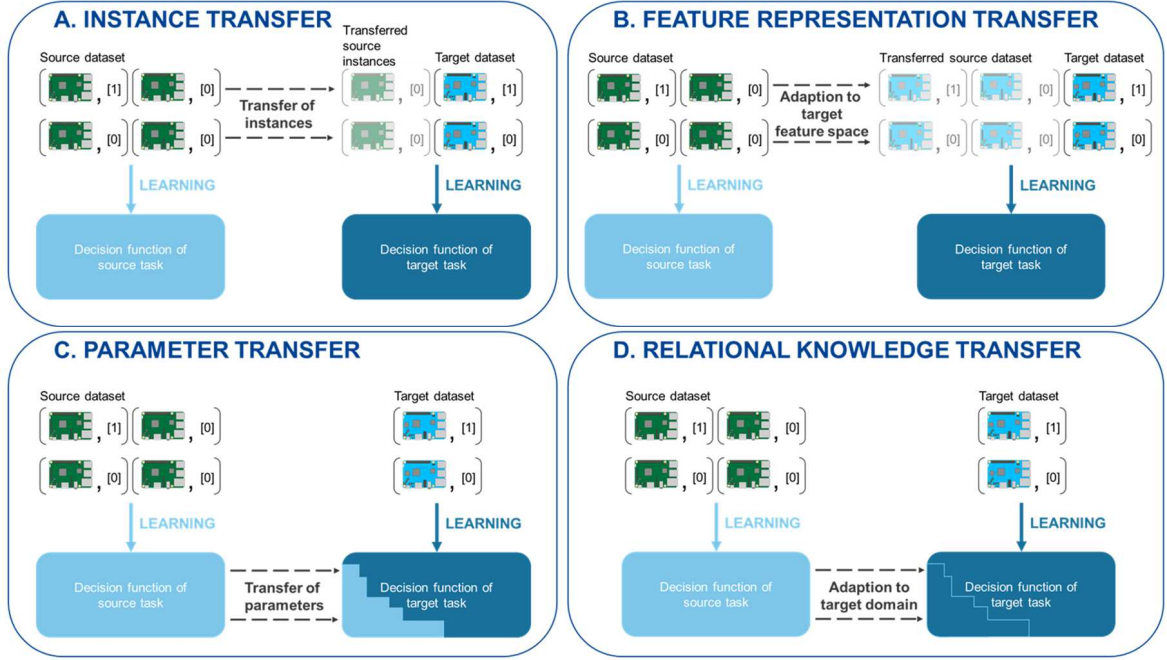

FIGURE 2. **Illustration of transfer learning approaches using the example of optical detection of manufacturing defects on printed circuit assemblies**

- **Relational Knowledge Transfer** (see Figure 2, Case D) describes approaches that directly map relational knowledge between source and target domains [8–10], typically requiring expert knowledge of those tasks' domains [9]. Using deep transfer learning, this can be alleviated by generative adversarial networks [15] or so-called end-to-end approaches that integrate domain adaption into the actual decision function [14].

While the first two categories represent primarily *data-driven* approaches, i.e. they rely on manipulating the datasets to convey the transfer, the latter two can be described as *model-driven* approaches, i.e. they significantly alter or exchange the models represented by the trained algorithms [10]. However, these terms are not to be mistaken for their meaning in other, more general contexts: Here, model-driven refers to abstract models representing the decision function, e.g. the weights and topology of a deep neural network, whereas e.g. mechanical, electrical or functional models are not being referred to.

It must be noted that the different approach categories' applicability to a given practical scenario does not primarily depend on their intrinsic (dis-)advantages in solving that problem in general nor the problem's categorization as introduced above, but rather the very specifics of the itself scenario, e.g. dataset and storage sizes, communication bandwidth or the availability of expert knowledge.

### 2.2 CONTINUAL LEARNING: TERMS AND DEFINITIONS

Although working on the transfer of knowledge for machine learning as well, the research on continual learning is largely independent of that on transfer learning. It has its own terminology and methodology, facilitating an interesting comparison with that of transfer learning.

While the term '*continual learning*' has been used in different contexts and various areas of research, here, it describes a field focused on solving multi-task classification problems with known source and target labels using deep learning techniques. For multi-tasking problems with identical label spaces, the term '*incremental learning*' or '*lifelong learning*' is used [12, 16]. For both, the goal is to sequentially learn tasks without forgetting previously learned tasks, an obstacle termed 'catastrophic forgetting' [17], so that eventually all tasks can be solved by a single deep learning algorithm [12, 16]. In continual learning, five different *problem categorization* classes can be differentiated:

- **Incremental Task Learning** describes problems, in which an algorithm trained to solve multiple (sub-)tasks with the same label space is able to infer an input's (sub-)task-specific label. For this purpose, the (sub-)task the input belongs to must be specified (see Figure 3) [18, 19].
- **Multi-Task Task Learning** describes problems similar to *incremental task learning* problems, but with differing label spaces [12].

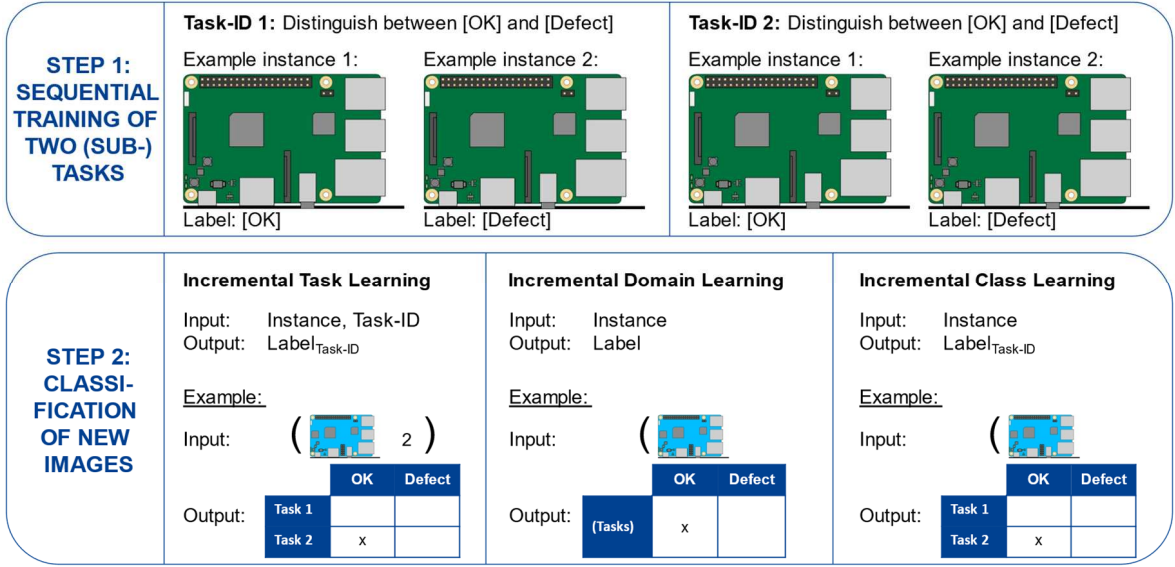

FIGURE 3.  **Illustration of continual learning problem categories using the example of optical detection of manufacturing defects on printed circuit assemblies**

- **Incremental Domain Learning** describes problems, in which an algorithm trained to solve multiple (sub-)tasks with the same label space is able to infer an input's abstract label. This abstract label is not associated with a specific (sub-)task (see Figure 3) [18, 19].
- **Incremental Class Learning** describes problems, in which an algorithm trained to solve multiple (sub-) tasks with the same label space is able to infer an input's (sub-)task-specific label (see Figure 3) [18, 19].
- **Multi-Task Class Learning** describes problems similar to *incremental class learning* problems, but with differing label spaces [12].

Compared with the *problem categorization* classes of transfer learning, continual learnings' *problem categorization* classes can be seen as sub-divisions of *homogeneous inductive learning*. However, rather than just adapting their skillset from solving one problem to another, continual learning approaches need to expand it to be capable of eventually solving all problems encountered [11, 12].

According to [16], three main *approach categorization* classes are commonly distinguished in continual learning. They can be associated with transfer learnings' *approach categorization* classes:

- **Architectural Strategy** describes approaches that change an algorithms architecture, e.g. its number or type of layers, in order to mitigate catastrophic forgetting. In transfer learning, they would be *parameter* or *relational knowledge transfer* approaches.
- **Regularization Strategy** describes approaches, that extend an algorithms loss function in a way that punishes changing weights important to solving previously learned tasks. In transfer learning, they would be *parameter transfer* approaches.
- **Rehearsal Strategy** describes approaches that periodically re-use data from previously trained tasks when training new tasks. In transfer learning, they would be *instance* or *feature representation transfer* approaches.

A more detailed discussion of those approaches can be found in [12, 16, 20] and a benchmarking of the most common of them regarding their performance in the three incremental learning problem classes can be found in [18, 19].

## 3. INDUSTRIAL APPLICATION OF DEEP TRANSFER LEARNING

While there are many publications focusing on the application of transfer learning in the fields of medical imaging, email spam or speech recognition [8, 10, 11, 14], to the authors' knowledge the industrial automation sector has not been surveyed in a similar fashion. Yet, in recent years, a growing number of deep transfer learning implementations tested and evaluated on industrial use cases has been published and shall therefore be exemplarily reviewed in this section. Building on this review, a novel categorization approach for industrial transfer learning use cases shall be introduced at its end.

TABLE 1. **Classification of implementations presented ('●' denotes a classification based upon scenarios and algorithms as described by the respective papers' authors, '○' denotes additional possible classes attributed upon analysis by this papers' authors)**

| | Problem categorization | | | | | | | | | Approach categorization | | | | | |
|---|---|---|---|---|---|---|---|---|---|---|---|---|---|---|---|
| | Base use case | | | | Label availability | | | Feature space similarity | | Transfer approach | | | | | Machine Learning Technique |
| | Cross-phase | Cross-state | Cross-entity | Cross-domain | Inductive | Transductive | Unsupervised | Homogenous | Heterogenous | Instance | Feature | Parameter | Relational | Continual | |
| *Anomaly detection* | | | | | | | | | | | | | | | |
| [21] | | ○ | ● | | ● | | | | | | | ● | | | Denoising AE |
| [22] | | ○ | ● | | | | ● | ● | | | | ● | | | LSTM AE |
| [6] | | ● | ○ | | ● | | | | ● | | | ● | | ○ | CNN + RNN |
| *Time series prediction* | | | | | | | | | | | | | | | |
| [23] | | ● | ○ | | ● | | | ● | | | | ● | | | LSTM |
| [7] | | ● | ○ | | - | | | | ● | | | | ● | | various |
| *Computer vision* | | | | | | | | | | | | | | | |
| [24] | | ● | ● | | - | | | ● | | | ● | | | ● | CNN + ARTMAP |
| [25] | | ● | ○ | | ● | | | ● | | | ● | | | | CNN |
| *Fault diagnosis* | | | | | | | | | | | | | | | |
| [26] | | ● | ○ | | ● | | | ● | | | | ● | | ○ | Sparse AE |
| [27] | | ○ | ○ | ● | ● | | | ● | | | | ● | | ○ | CNN |
| [28] | | ● | ○ | | ● | | | ● | | | | ● | | ○ | CNN |
| [29] | ● | | | | ● | | | ● | | | | ● | | ○ | Stacked Sparse AE |
| [30] | | ○ | ○ | ● | ● | | | ● | | | ● | | | | CNN |
| [31] | | ○ | ○ | ● | ● | | | ● | | | | ● | | ○ | Adversarial CNNs |
| *Fault prognostics* | | | | | | | | | | | | | | | |
| [32] | | ○ | ● | | | | ● | ● | | | | ● | | ○ | Sparse AE |
| *Quality Management* | | | | | | | | | | | | | | | |
| [33] | ● | | | | ● | | | ● | | | | ● | | | MLP |
| [34] | | | | ● | ● | | | ● | | | | ● | | | Stacked CNN |

The different implementations presented are categorized in Table 2. Because of a lack of comparability between the respective scenarios or use cases, listing and comparing resulting performance metrics (e.g. absolute accuracy, gain compared to baseline approaches) is not feasible. Instead, a more holistic assessment of each reviewed paper's methodology is given at the end of each sub-section.

### 3.1 ANOMALY DETECTION

*Anomaly detection* aims at finding data samples that differ from the majority of samples in a dataset. This classification task is the basis for many other applications, e.g. fault detection. Deep learning methods are well suited for anomaly detection [35] and some generic (i.e. focusing just on anomaly detection without any more specific application) implementations of deep transfer learning have been published:

*Parameter transfer* is used in [21] to transfer knowledge between two aluminum extrusion electricity consumption datasets. A denoising auto-encoder (AE) is pre-trained unsupervisedly on a large source dataset and then fine-tuned on the target dataset.

In [22], *parameter transfer* improves anomaly detection on simple process sequences across different production chambers. Here, the last encoding and decoding layer of a pretrained long short-term memory (LSTM)-based AE are re-trained on the target dataset.

Similarly, using *parameter transfer*, [6] allows for changes in the feature space caused by different numbers and types of sensors in an elevator monitoring scenario without needing to retrain the complete algorithm. This implementation relies on one-dimensional convolutional neural networks (CNN) for each sensor to extract features, which are then aggregated and classified by a recurrent neural network (RNN).

While [21] and [22] show better performances than conventional deep learning algorithms (DLA), their informative value is limited: [22]'s process sequence dataset is far from realistic and the comparison carried out in [21] uses an overly simple, single-layer artificial neural network. [6], however, is methodologically very thorough, analyzing multiple distributions of 'normal' and 'anomalous' samples, different RNN algorithms and anomaly types (point, context-specific and collective) on a very relevant use case. While not outperforming its conventional competition in all scenario and algorithm variations, its detailed analysis should facilitate good adaptability.

### 3.2 TIME SERIES PREDICTION

*Time series prediction* aims at predicting non-linear time-variant system outputs based upon time series data. This regression task is the basis for many other applications, e.g. fault prediction. It is commonly executed by deep learning methods [36] and some generic (i.e. focusing just on time series prediction without any more specific application) implementations of deep transfer learning have been published:

*Parameter transfer* is used in [23] to transfer knowledge between two vibrational datasets acquired from aircraft engines on the ground and in flight. An LSTM is pre-trained on the source dataset and then fine-tuned on the target dataset. Based on its title, the authors would have expected this paper to focus more on "anomaly detection".

In [7], *relational knowledge transfer* is used to transfer knowledge among different generic machine datasets. Those datasets are assigned parameters by an expert that allow for a transfer without the need to retrain on target data at all. This is achieved by mapping the different feature spaces into a common one in which the actual regression can be carried out by a variety of algorithms including RNNs.

Compared with training on the target data from scratch, [23] achieves a lower prediction error while needing less training data. Furthermore, [23] is a good example for the need for a coherent terminology, as the transfer learning terms used therein greatly differ from the definitions in the papers presented above. The approach presented in [7] is thoroughly analyzed with comparisons between different algorithms and transfer scenarios. Not all perform better using transfer learning, but good reasoning is presented for the ones that do not.

### 3.3 COMPUTER VISION

*Computer vision* aims at extracting complex information out of image data. This classification task is a core strength of CNNs and commonly addressed using a *parameter transfer* approach: A CNN is pre-trained on a large, standard dataset of natural images. Only its last layer(s) are then fine-tuned to the task at hand in order to reduce resource requirements for training [5]. Still, some implementations use transfer learning approaches to realize benefits beyond that:

*Feature representation transfer* is used in [24] to transfer knowledge among different image datasets. A lightweight CNN condenses images from a life image stream to feature vectors, which are then classified by a fuzzy ARTMAP algorithm. The class representations can be transferred to other entities of the learning algorithm allowing a rapid exchange of learnt knowledge.

To transfer knowledge acquired from depth to corresponding color images, [25] relies on *representational knowledge transfer*. In this cross-modal approach, a CNN pre-trained on images from the depth image dataset known to deliver good results is further trained on the corresponding color images to output the same continuous grasp quality score.

Compared to other approaches, [24] reduces the amount of computing power and storage required to solve its distributed, incremental learning task. Its problem being regression not classification, [25] thoroughly analyzes different transfer settings which all show large improvements compared with the baseline non-transfer approach. However, the selection of the right training images facilitating the transfer needs to be carried out manually leaving room for further improvement. Yet, the evaluation is carried out using an actual robot arm to grasp the objects represented in the images and good reasoning is presented for faulty grasp attempts.

### 3.4 FAULT DIAGNOSIS

*Fault diagnosis* aims at classifying fault events in a dataset. This classification task is commonly carried out by deep learning methods [2] and a considerable number of deep transfer learning implementations has been published:

*Relational knowledge transfer* in an end-to-end domain adaption approach is used by several implementations [26–28] to transfer knowledge between different operational states within one or more bearing vibration datasets or between different such datasets – among them always the Case Western Reserve University's Bearing Dataset (CWRUBD). While [26] uses a sparse AE, [27] and [28] use CNNs trained on the source and target datasets to extract features into a common feature space where they are classified by a simple classification layer trained on the labels of the source dataset.

*Relational knowledge transfer* is used in [29] to transfer knowledge between two programmable logic controller datasets – one from a real, one from a simulated version of the same shop floor. A stacked sparse AE is pre-trained on the (simulated) source dataset to extract its features. Enhanced by an adaption layer, it subsequently fits both domains when training continues on source and target datasets simultaneously. The features are then classified by a simple, continuously adapting classifier.

Using vibration data from the CWRUBD, again, and two other, self-created rotational wear datasets, [30] relies on *parameter transfer*. The top-most layers of a CNN trained on standard natural images are adapted to the fault classification task using the source dataset. The actual in-scenario transfer is then carried out by fine-tuning these top-most layers using the target dataset.

*Relational knowledge transfer* is used in [31] to transfer knowledge between vibration data from four different bearing datasets – among them CWRUBD. In a domain adversarial network algorithm, a CNN-based domain discriminator incentivizes a CNN-based feature extractor to extract features from both source and target domains into a common feature space where they are classified.

While performing slightly better than a conventional DLA, the close similarity between the different sub-datasets greatly limits the informative value of [26]. Suffering the same problem regarding the CWRUBD dataset, [30] can show larger performance and training speed gains on its other datasets. [27, 28, 31] all analyze their proposed algorithms thoroughly, showing considerably better results while taking less training time than comparable approaches. While all those implementations were evaluated on vibrational data, the broad variety of approaches should allow for good overall adaptability to other data types. Quite different from the other implementations, [29], too, shows good results. Unfortunately, the characteristics of the dataset used for evaluation should have been described more in detail.

### 3.5 FAULT PROGNOSTICS

*Fault prognostics* aims at predicting the remaining useful lifetime (RUL), i.e. the time to failure, of an entity. This regression task is increasingly carried out by data-driven methods, as well [2]. Deep transfer learning, however, has not been extensively researched yet:

*Relational knowledge transfer* is used in [32] to transfer knowledge from one vibration dataset characterizing a CNC tool to another. In this end-to-end approach, a sparse AE trained on the source and target datasets is used to extract features into a common feature space where a non-linear regression layer trained on the source dataset predicts the RUL value.

The approach used shows good results compared with a conventional DLA. However, the evaluation scenario appears to be rather limited, calling for broader examination. Still, the simplicity of the approach renders it a good candidate for adaption.

### 3.6 QUALITY MANAGEMENT

*Quality management* aims at generating quality metrics for production processes. It can range from the retroactive detection of faulty products to the proactive prediction of faults based upon current parameters, encompass regression as well as classification tasks. It is the basis for counter measures termed predictive or autonomous quality control [37]. Depending on the specific scenario, a variety of approaches exist – among them some involving deep transfer learning:

*Parameter transfer* is used in [33] to transfer knowledge from a simulated injection modeling process to a real one. Two approaches are examined: A simple multi-layer perceptron (MLP) is trained first on the source dataset and then just fine-tuned on the target dataset ('soft-start') or its hidden layers are completely trained anew ('random initialization').

Using *parameter transfer*, too, [34] addresses the knowledge transfer between an out-of-domain natural image dataset and a metal casting x-ray image dataset. A modular CNN architecture for feature extraction, region of interest proposal and classification is first trained on the source dataset. Then it is fine-tuned regarding feature extraction and region of interest proposal while being re-trained after resizing for the classification part.

Conducting quality prediction on a continuous scale, [33] achieves good results even on very small datasets using its soft-start approach. Contrastingly, [34] focusses on retroactive quality monitoring, out-performing conventional DLA, too.

### *3.7 BASE USE CASES FOR INDUSTRIAL TRANSFER LEARNING*

Looking at the review presented in Table 2 and the aforementioned concepts of knowledge transfer, it becomes apparent that neither the continual learning nor the transfer learning problem categorization sufficiently reflect the motivations underlying the industrial application of such techniques. Instead of the focus on abstract characteristics such as label availability or domain similarity, a more use case oriented categorization is needed to structure the examination of best practice examples and facilitate their adaption to other scenarios. The authors therefore propose the following four base use cases motivating industrial transfer learning:

A. **Cross-phase** industrial transfer learning refers to the transfer of knowledge from one lifecycle phase to another, e.g. from engineering to operations (see Figure 4, Case A).
B. **Cross-environment** industrial transfer learning refers to the transfer of knowledge from one process to a similar process. It can be further sub-divided regarding the nature of this (dis-) similarity (see Figure 4, Case B):
    B.I. **Cross-state** industrial transfer learning refers to the transfer of knowledge within one lifecycle phase of one entity, e.g. through dynamically changing operational states of this entity (see Figure 4, Case B.I).
    B.II. **Cross-entity** or collaborative industrial transfer learning refers to the transfer of knowledge from one location to another, e.g. across different sites where entities of the same type are used or just from one entity to another (see Figure 4, Case B.II).
C. **Cross-domain** industrial transfer learning refers to the transfer of knowledge from one distinct domain or process to another, e.g. from the production of one product to the production of a different product (see Figure 4, Case C).

In practice, these four base use cases for industrial transfer learning often overlap: On the one hand, scenarios can incorporate aspects of several base use cases (e.g. [24]). This effect increases with growing scenario complexity or scenario realism. On the other hand, one algorithm can be capable of solving several different base use cases. This becomes apparent for cross-environment use cases in Table 2, where with all cross-state algorithms (e.g. [6], [7], [25]) described by the respective authors cross-entity scenarios would be solvable, too – and vice-versa (e.g. [21], [22], [32]). Moreover, most reviewed cross-domain algorithms ([27], [30], [31]) are capable of solving both cross-environment use cases as well. However, this does not mark the proposed classes redundant or overly specific, but merely reflects the different perspectives and needs of practitioners describing use case characteristics and scientists classifying advanced research.

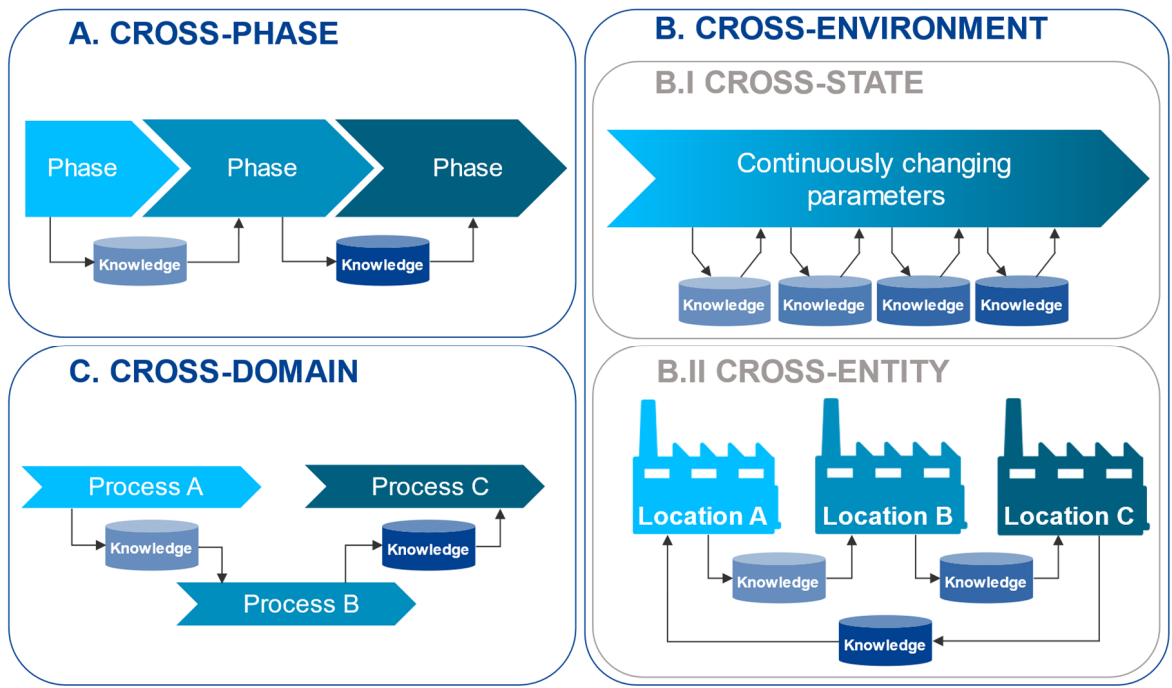

FIGURE 3. **Base use cases for industrial transfer learning**

The proposed base use cases for industrial transfer learning thereby promise to provide a valuable reference structure for describing and comparing areas of application, adding another dimension to the already established categories for problem and approach categorization.

## 4. DISCUSSION

While the exemplary nature of the review of industrial applications of deep transfer learning conducted above does not support a quantitative analysis, some trends and general insights can be deduced, nonetheless.

### *4.1 PROBLEM PERSPECTIVE*

Depending on the specific use case, deep transfer learning is no novelty, but state-of-the-art: CNN-based computer vision and image manipulation algorithms have been relying on parameter transfer in order to perform well on edge devices for years. Yet, in most other problem classes, the use of deep transfer learning truly just started [14].

One reason for this discrepancy – despite compelling arguments for deep transfer learning presented by the publications listed above – might be the lack of large, standardized and publicly available non-image datasets [14]. It is no surprise that almost all fault diagnosis examples use the CWRUBD and many fault prognostics publications (not only the one covering a transfer learning approach) use the turbofan engine dataset. Public datasets allow for a direct comparison of different methodologies and such comparison drives scientific advancement.

Still, the presented implementations provide examples for solving a large variety of different problem types, among them regression as well as classification tasks and all four proposed base use cases of industrial transfer learning. While usually only explicitly addressing one of the latter, many approaches are capable of more than that, e.g. most methods solving cross-state problems can fundamentally solve cross-entity problems as well – and vice versa (see '○' markings in Table 2).

### *4.2 APPROACH PERSPECTIVE*

Based on Table 2, data-driven transfer approaches appear to be less frequently used than parameter and relational knowledge transfer. This might be caused by the need for transferable instances to be selected based upon expert knowledge in instance transfer and the lower degree of automation in feature representation transfer compared with the end-to-end approach in representational knowledge transfer [14].

Contrastingly, parameter transfer, oftentimes implemented as a mere finetuning of an algorithm pre-trained on the source dataset(s), is easy to implement and well documented, e.g. in the many cases of pre-trained CNN-based computer vision algorithms. More sophisticated and thereby perhaps showing better performance, relational knowledge transfer is frequently used as well.

The separation of *continual* from *transfer learning* is very much present with only one publication explicitly mentioning the former. This is unfortunate, as transfer learning tends to facilitate differentiation among different versions of an algorithm while continual learning leads to an algorithm's generalization [38] – and only a combination of both features will truly enable the robust transfer of knowledge across all base use cases as introduced before. Just focusing on transfer learning to solve industrial automation use cases will therefore not suffice. This notion is underlined by authors even from the field of transfer learning arguing for *online* or *sequential* learning capabilities very similar to *incremental learning*'s learning on expanding datasets as a necessary enabler for a more widespread practical utilization of deep learning [11, 12, 14, 39].

Yet, a combination of continual and transfer learning approaches is not the only technological merger that could greatly enhance the adoption of industrial transfer learning. There is a strong overlap between cross-phase industrial transfer learning approaches and the concept of *intelligent digital twins* referring to artificially intelligent, synchronized, digital representations of physical assets actively acquiring new data and being simulable [40]: The lifecycle-spanning digital twin requires the ability to transfer knowledge learnt across any changes the physical asset might encounter. Cross-phase industrial transfer learning addresses this problem but requires a close synchronization between simulation and reality as well as a way to store data from one lifecycle-phase to another. While first implementations work on joining both technologies [29, 33, 41], there still is much potential for further advancement.

## 5. CONCLUSION

In this article, the concepts of transfer learning, focusing on using source task knowledge to benefit a target task, and continual learning, focusing on using source and target task knowledge to benefit both tasks, are laid out. From the ensuing review of implementations in industrial automation, it becomes apparent that this abstract differentiation is not benefitting the practical utilization of neither class of algorithms.

In contrast, from a practical perspective, both classes of algorithms should be brought together to create robust learning algorithms fulfilling the requirements of the industrial automation sector. To better describe these requirements, base use cases of industrial transfer learning are introduced.

The review further revealed a number of promising approaches and best practice examples for industrial deep transfer learning – some even implicitly bridging the aforementioned gap between continual and transfer learning by showing traits of both. However, many of the implementations discussed were evaluated on simple, abstract tasks only loosely resembling real world industrial automation challenges. Here, more elaborate tasks and corresponding datasets– open access, if possible – are needed to better assess and possibly compare the applicability of presented approaches. Furthermore, focus should be laid on regression tasks capable of handling multi-variate time series data as they represent a large percentage of industrial automation tasks and are underrepresented in many other fields. The examples presented therefore only scratch the surface of possible implementations and applications, calling for more research in this highly dynamic field.